\documentclass[letterpaper]{article} 
\usepackage[draft]{arxiv}  
\usepackage{times}  
\usepackage{helvet}  
\usepackage{courier}  
\usepackage[hyphens]{url}  
\usepackage{graphicx} 
\usepackage{natbib}  
\usepackage{caption} 
\usepackage{multirow}
\usepackage{amsmath}
\usepackage{booktabs}
\usepackage{algorithm}
\usepackage{algorithmic}
\usepackage{amsfonts}
\usepackage{xcolor}

\usepackage{newfloat}
\usepackage{listings}
\usepackage{amssymb}
\usepackage{cuted}  
\newcommand{\myparagraph}[1]
{\vspace{0.1em}\noindent\textbf{#1}}
\usepackage{pgfplots}
\pgfplotsset{compat=1.18}
\DeclareCaptionStyle{ruled}{labelfont=normalfont,labelsep=colon,strut=off} 
\floatstyle{ruled}
\newfloat{listing}{tb}{lst}{}
\floatname{listing}{Listing}
\title{SCOPE: Sign Language Contextual Processing with Embedding from LLMs}
\author{
    Yuqi Liu\thanks{The authors contributed equally},
    Wenqian Zhang\footnotemark[1],
    Sihan Ren,
    Chengyu Huang,
    Jingyi Yu,
    Lan Xu\thanks{Corresponding author}
}
\affiliations{
    School of Information Science and Technology, ShanghaiTech University\\
    \{liuyq2, zhangwq2022, rensh2022, huangchy, yujingyi, xulan1\}@shanghaitech.edu.cn
}

\usepackage{bibentry}

\begin{document}

\maketitle

\begin{strip}
    \centering
    \includegraphics[width=1.0\linewidth]{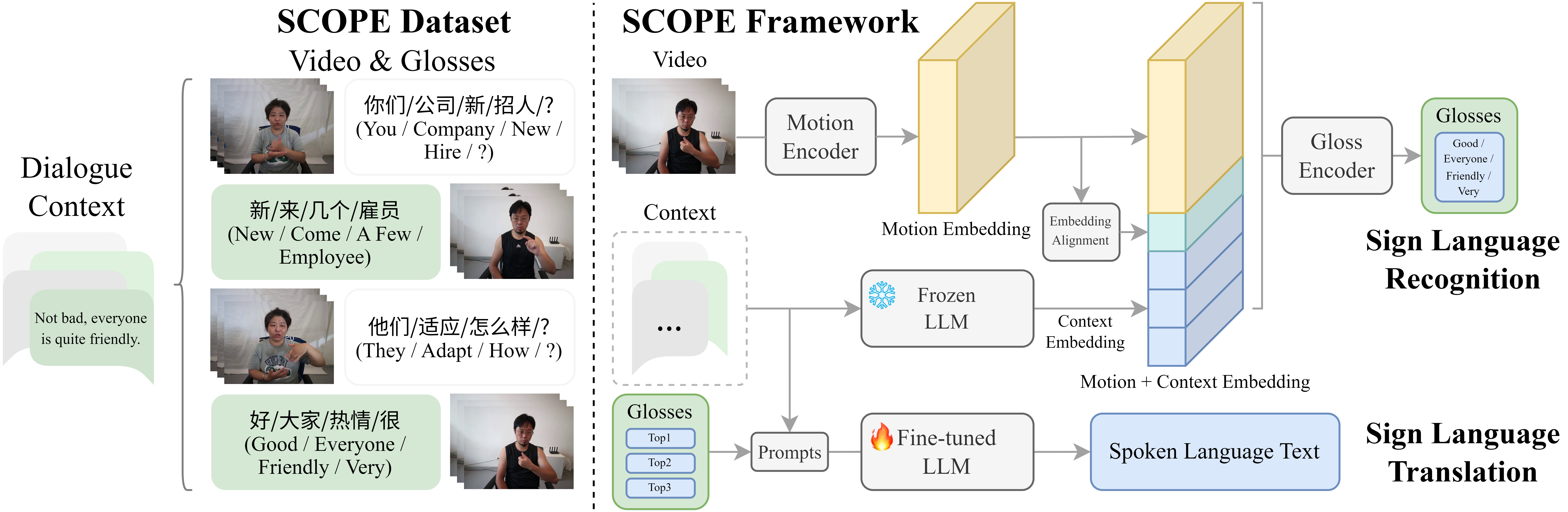}
    \captionof{figure}{(a) Our SCOPE dataset contains rich contextual information and sign language videos. (b) Our SCOPE framework is a robust context-aware sign language recognition/translation model capable of recognizing dialogue-based sign language gestures, predicting glosses, and generating spoken sentences with the aid of LLMs.}
    \label{img:teaser}
\end{strip}

\begin{abstract}
Sign languages, used by around 70 million Deaf individuals globally, are visual languages that convey visual and contextual information. Current methods in vision-based sign language recognition (SLR) and translation (SLT) struggle with dialogue scenes due to limited dataset diversity and the neglect of contextually relevant information.
To address these challenges, we introduce SCOPE (Sign language Contextual Processing with Embedding from LLMs), a novel context-aware vision-based SLR and SLT framework. For SLR, we utilize dialogue contexts through a multi-modal encoder to enhance gloss-level recognition. For subsequent SLT, we further fine-tune a Large Language Model (LLM) by incorporating prior conversational context. 
We also contribute a new sign language dataset that contains 72 hours of Chinese sign language videos in contextual dialogues across various scenarios. 
Experimental results demonstrate that our SCOPE framework achieves state-of-the-art performance on multiple datasets, including Phoenix-2014T, CSL-Daily, and our SCOPE dataset. 
Moreover, surveys conducted with participants from the Deaf community further validate the robustness and effectiveness of our approach in real-world applications. 
Both our dataset and code will be open-sourced to facilitate further research.

\end{abstract}

\section{Introduction}

Sign language is the vital visual language used by the Deaf and hard of hearing. Hence, vision-based sign language understanding provides a communication bridge between the Deaf and hearing communities. Such a bridge should accurately and conveniently convey complex contextual information during communication between us humans, especially for dialogue scenarios.

Currently, the two main tasks in vision-based sign language processing include Sign Language Recognition (SLR)~\cite{jiao2023cosign,wei2023improving,zheng2023cvt} and Sign Language Translation (SLT)~\cite{zhao2024conditional,chen2022two,yin2023gloss}. SLR converts visual signals into intermediate gloss sequences~\cite{stokoe2005sign}, while SLT translates visual signals or glosses into natural language. Yet, we notice that most existing methods, both SLR and SLT, focus on translating one sentence at a time, largely ignoring the contextual information of dialogue scenes.
Indeed, it's mainly due to the severe lack of sign language datasets for dialogue scenes with sufficient contextual information. For example, the widely adopted PHOENIX2014\cite{koller2015continuous} and PHOENIX2014T datasets\cite{camgoz2018neural} focus on weather forecasts. The How2Sign dataset\cite{duarte2021how2sign} addresses everyday scenarios but only contains isolated sign language sentences. The CSL-Daily dataset\cite{zhou2021improving} contains daily sentences but they lack preceding or following context and are essentially still independent statements. 
On the other hand, in the field of Natural Language Processing,  recent advances~\cite{ouyang2022training,touvron2023llama,qwen2} with large language models (LLMs) have demonstrated that contextual information can significantly improve semantic understanding and linguistic abilities. Some recent methods~\cite{gong2024llms,wong2024sign2gpt} utilize LLMs for sign language understanding. Yet, they still focus on per-sentence translation and fall short of analyzing the contextual information for dialogue scenarios. In a nutshell, both the dataset and methodology for contextual vision-based sign language processing remain far-reaching.

To this end, we introduce \textit{SCOPE}, a contextual sign language recognition and translation approach tailored for the dialogue scenes, as shown in Fig.~\ref{img:teaser} for overview. Specifically, we first contribute a context-based dataset of Chinese sign language dialogues. Our dataset covers a wide range of both daily and professional conversations like shopping and medical treatment. It includes 59,231 dialogue sequences totaling 72.4 hours. For each sequence, we provide video footage, gloss annotations, and dialogue texts, all by professional Deaf individuals from diverse backgrounds.

Secondly, we provide a strong baseline for vision-based contextual sign language processing, which organically utilizes recent LLMs to extract the contextual information from our unique dataset.
For the SLR task, we extract sign motion features from the video footage and then introduce a novel embedding alignment module to align them to the context embeddings from a frozen LLM. Then, we feed these aligned motion/context embeddings into a gloss encoder to obtain the recognized gloss sequences. We observe that such an alignment between the current motions and the preceding contextual information from the LLM is crucial for performance gain. It preserves both the motion and semantic information of the sign language while enabling the concatenation of the contextual embeddings with the input. For the subsequent SLT task, we further leverage the contextual understanding capabilities of the LLM. We use the gloss output from the previous SLR module and the contextual text as inputs and adopt Q-LoRA~\cite{dettmers2023qlora} to efficiently fine-tune a pretrained LLM model, achieving accurate and natural translations that are closely aligned with the context.

For validation, we conduct comprehensive experiments on both our unique contextual dataset and previously context-free datasets and showcase a companion live demo for sign language translation, which demonstrates the state-of-the-art performance of our approach.
In summary, we provide a novel vision-based, context-driven sign language processing approach that utilizes LLMs to address SLR and SLT tasks in dialogue and communication settings. We also contribute a large-scale contextual dataset of Chinese sign dialogues. 
We believe that both our dataset and baseline approach are the first of their kind to open up the research direction towards context-aware and vision-based sign language analysis. Both our benchmark dataset and baseline approach will be made publicly available.

\section{Related Works}
\begin{figure*}[t]
	\centering
	\includegraphics[width=1.0\linewidth]{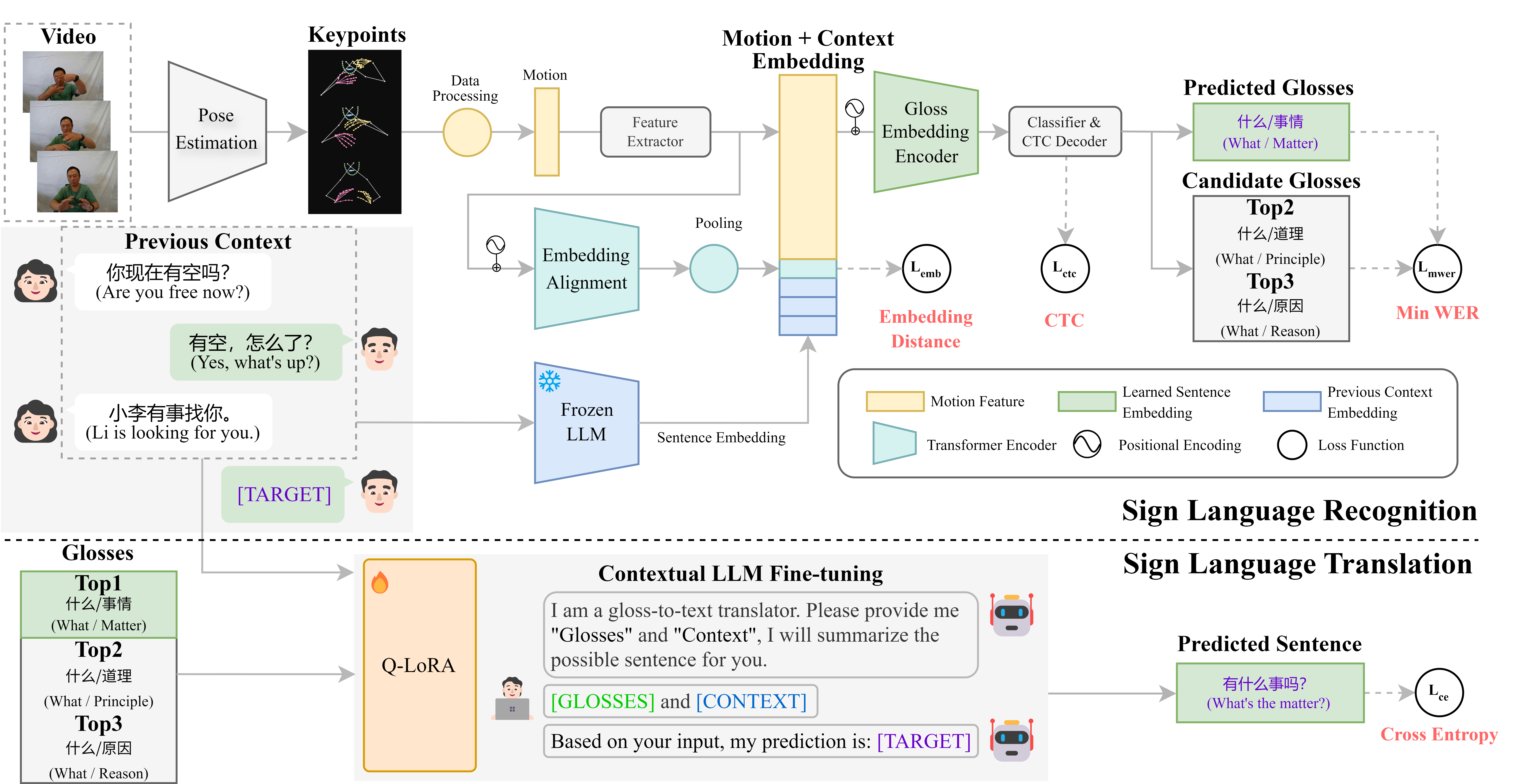}
	\caption{\textbf{Overview of SCOPE framework.} 
Our Embedding Alignment Encoder captures holistic linguistic information from the whole motion sequence.
Aligning embedding space to match a frozen LLM enables integrating previous context information for SLR.
Finally, Q-LoRA fine-tuning fits an LLM for translating predicted glosses with context into spoken language.} 
	\label{img:method}
	\vspace{-5mm}
\end{figure*}
\myparagraph{Sign Language Recognition}
(SLR) focuses on recognizing glosses from sign videos. While progress has been made in Isolated SLR (ISLR) \cite{albanie2020bsl, tunga2020posebasedsignlanguagerecognition, li2020transferring, hu2021signbert, li2020word}, current research is shifting to Continuous SLR (CSLR), which converts continuous sign videos into sentence-level gloss sequences. This task involves two main components: feature extraction and translating these features into gloss sequences.


Visual feature extraction often involves extracting features from RGB images using CNNs\cite{chen2022simple, li2020tspnethierarchicalfeaturelearning, hu2023self, min2021visual}. These features are then modeled with temporal frameworks like RNNs\cite{camgoz2018neural, ko2019neural}, LSTMs\cite{hu2023continuoussignlanguagerecognition, cui2019deep}, and Transformers\cite{camgoz2020sign, voskou2021stochastic, yin2020better} to capture the connection between visual signals and glosses. Some approaches\cite{zhou2021spatial, papadimitriou2020multimodal} utilize estimated keypoint sequences to describe motions through spatial coordinates or generate heatmaps\cite{chen2022two, chen2024signvtclmultimodalcontinuoussign}. However, many methods require video processing, which can be slow and space-consuming, limiting their practical application.

Decoding the extracted features into gloss sequences needs temporal modeling. Hidden Markov Models (HMMs)\cite{8099847, GAO20042389, koller2016deep} and Connectionist Temporal Classification (CTC)\cite{cheng2020fully, zhou2021spatial, min2021visual} are commonly used for this purpose. However, most existing methods focus on frame-wise or sentence-wise information, often neglecting the broader linguistic context, resulting in the loss of important language features.

\myparagraph{Sign Language Translation}
(SLT) aims to translate sign language directly into natural language, bridging the gap between the Deaf community and hearing individuals. This task is challenging due to the modality gap between visual signal and text, compounded by the scarcity of context sign language datasets. Many approaches\cite{camgoz2020sign, zhou2021spatial, zhou2021improving} use SLR results to aid translation. Joint training of SLR and SLT modules has also been explored to improve performance. Some researchers\cite{li2020tspnethierarchicalfeaturelearning, camgoz2018neural, zhou2023glossfreesignlanguagetranslation} seek to eliminate gloss by directly translating sign language videos into text using techniques like Conditional Variational Autoencoders and Transformers. SLT involves projecting visual features into coherent textual representations, necessitating insights from both computer vision and natural language processing. Key advancements leverage pretrained language models like mBART\cite{liu2020multilingual} for enhanced textual understanding\cite{chen2022two, chen2022simple}. Recent studies also explore the use of frozen and fine-tuned large language models\cite{wong2024sign2gpt, gong2024llms} to improve translation quality.

\myparagraph{Sign Language Dataset.}
Progress in sign language research has been driven by data. Many researchers have contributed valuable datasets of isolated signs \cite{wang2016isolated, zhang2016chinese, joze2018ms, imashev-etal-2020-dataset, sridhar2020include, li2020word, AUTSL, albanie2020bsl, desai2024asl}. However, while each video clip corresponds to a single sign, the practical utility of such data remains limited.

There are several recent datasets that provide continuous sign language data. For instance, the PHOENIX-2014\cite{koller2015continuous} dataset includes sign language videos from television broadcasts along with corresponding gloss annotations, primarily focusing on weather forecasts. Datasets like SIGNUM\cite{von2008}, PHOENIX-2014T\cite{camgoz2018neural}, and CSL-Daily\cite{zhou2021improving} not only offer gloss annotations but also include natural language translations of the signs, thereby advancing Sign Language Translation (SLT) research. Additionally, the CCSL\cite{huang2018video} dataset provides images with depth information, increasing the information of sign data. The How2Sign\cite{duarte2021how2sign} dataset stands out with its multi-view information, enabling the capture of 3D sign language motions.

Despite improvements in the size and diversity of sign language datasets, they remain limited in domain coverage. Current corpora consist of context-independent sentences, lacking the contextual relationships needed to fully utilize the linguistic features of sign language, which hinders advancements in SLT research.

\section{Method}
We present SCOPE framework, a novel framework that aligns motion features with LLM-provided
sentence embeddings of previous contexts, aiming to fully utilize contextually related dialogues in which sign language conversations mainly occur. To address the often overlooked contextual aspects in data collection, we provide SCOPE dataset that annotates sign videos with additional context information, which our model effectively utilizes. Details of SCOPE dataset will be further presented in the Dataset section.

Fig. \ref{img:method} demonstrates the structure of our SCOPE framework. 
Our Embedding Alignment Encoder transforms motion features into an embedding that captures the linguistic information of the whole motion sequence. 
Aligning embedding space to a frozen LLM enables integrating contextual information of previous sentences to recognize glosses.  
Finally, Q-LoRA fine-tuning fits an LLM for translating predicted glosses into spoken language with the assistance of context information. 

\subsection{Model Details}\label{seq:slr}
\myparagraph{Embedding Alignment Encoder.}
We use a transformer encoder structure to extract information from motion features. For the input keypoints $\textbf{J}={J_{1}...J_{t}}$, they first pass through the feature extractor linear layer and the temporal sequencer linear layer, which compress the motion information in the spatial and temporal dimensions, respectively, resulting in the intermediate state motion input $D$, which aligns textual embedding in shape. 

Next, we need to pretrain an Embedding Alignment Encoder to align features from the motion space with the textual embedding space. The key idea is to directly learn the alignment between the linguistic features of sign language motion and the contextual features of text. In this step, we aim to align the sign motion feature $D$ with the embedding vector of the target sentence. We do this by passing the motion features through the Embedding Alignment transformer encoder and then pooling them to compress the time dimension, resulting in an embedding vector that matches the size of the text embedding. The encoding process, in detail, first embeds the input $D$ into a latent space, represented by $h_{0}$, and then obtains the encoded hidden states $h_{n}$ through $N$ attention layers. Finally, a feed-forward network is used to obtain the encoded vector. The formulas for the transformer motion encoder process are as follows:
\begin{equation} \label{eq:motiontransformer}
\begin{split}
Q &= W^{Q}h_{i}, K = W^{K}h_{i},V = W^{V}h_{i},\\
h_{i+1} &= \textit{Attn}(Q,K,V) = \textit{softmax}(\frac{QK^{T}}{\sqrt{C}})V,
\end{split}
\end{equation}
where $W^{Q}$,$W^{K}$,$W^{V}$ are trainable weights, $C$ is the number of channels in the attention layer, and $h_{i+1}$ is hidden states before the next layer.

Supervision by distance to the embedding of the target sentence provided by an LLM.

The loss of the motion encoder is the L2 distance between the pooled embedding vector and the target text embedding vector.
\begin{equation} \label{eq:embedding loss}
\mathcal{L}_\textit{emb} = \mathbb{E}||E_\textit{out}-E_\textit{text}||_{2},
\end{equation}
where $L_{\text{emb}}$ denotes the embedding loss, $E_{\text{out}}$ is the output of the motion encoder, and $E_{\text{text}}$ is the text embedding vector. The text embedding vector is generated by a frozen LLM text embedding model \cite{neelakantan2022text}, which encodes the ground truth sentence meaning of the sign video. Through this process, we align the motion features with the language feature information, enhancing the connections between the isolated sign words.

\myparagraph{Gloss Embedding encoder.}
After aligning the motion features, we obtain an embedding vector that contains both semantic and sign language information. Next, we combine this with the motion features to predict the gloss. For sign language conversations, providing previous language context is crucial for improving the accuracy of recognizing the current target sentence. Therefore, we use a frozen LLM to get the embedding vector for the previous sentences. To minimize irrelevant information, we only keep the last three text embeddings. If there are fewer than three previous sentences, we use a mask to ignore the padding input. The gloss embedding encoder is also a transformer encoder model. The encoding process can be formulated as follows:
\begin{equation} \label{eq:embeddingtransformer}
\begin{split}
H^{0}_{t,A} &= \textit{Hidden}(\textit{Cat}(E_{t},E_{A})), \\
E_\textit{out} &= \textit{FFN}(\textit{Attn}(W^{Q}H^{N}_{t,A},W^{K}H^{N}_{t,A},W^{V}H^{N}_{t,A})),
\end{split}
\end{equation}
where $\textit{Hidden}$ is the hidden layer embedding in the transformer encoder, and $\textit{Cat}$ is the concatenate operation. $E_{t}$ is the previous stage encoded sequence, and $E{A}$ is the above text embedding vector. $\textit{FFN}$ is the feed-forward network in the transformer encoder. Passing the output $E_\textit{out}$ through a linear classifier layer, we get the output logic of the glosses. 

\myparagraph{CTC Decoding.} We use connectionist temporal classification (CTC) \cite{graves2006connectionist} loss to optimize the embedding encoder: 
\begin{equation} \label{eq:ctcloss}
\mathcal{L}_\textit{CTC}^{y} = -\textit{log}p(\textbf{l}|y) = -\textit{log}_{\pi \in \mathcal{B}^{-1}(\textbf{l})}p(\pi | y) , 
\end{equation}
where $\textbf{l}={l_{1}...l_{t}}$ is the gloss sequence corresponds to keypoints sequence $J$. $\mathcal{B}$ is a many-to-one mapping between hypotheses and gloss, and $\pi$ is the alignment path. 
In addition, we adopt Minimum World Error Rate (MWER) Training \cite{meng2021minimum} technique to reduce the mismatch between training objectives and evaluation metrics, boosting the accuracy of hypotheses on top of the beam. We use beam search during training to decode the top 3 possible gloss sequences. While maintaining the top 1 decoded result as the final output of the SLR network, other candidate glosses contribute to optimization with minimum word error rate (MWER) loss:
\begin{equation} \label{eq:minwerloss}
\mathcal{L}_\textit{MWER}=\sum_{n=1}^N \bar{P}(\mathbf{Y}^n | \mathbf{J};\theta)R(\mathbf{Y}^n, \mathbf{Y}^*),
\end{equation}
 where $\bar{P}(\mathbf{Y}^n | \mathbf{J};\theta) = \frac{P(\mathbf{Y}^n | \mathbf{J};\theta)}{\sum_{n=1}^N P(\mathbf{Y}^n | \mathbf{J};\theta)}$, is the re-normalized posterior over the N-best hypotheses, $\theta$ is model parameters, and $R(\mathbf{Y}^n, \mathbf{Y}^*)$ is the number of word errors in a hypothesis $\mathbf{Y}^n$ compared to the reference $\mathbf{Y}^*$.
 
Furthermore, the top 3 decoded results also serve as the input of the LLM model in SLT task.

\myparagraph{Contextual LLM Fine-tuning.} Inspired by \cite{gong2024llms}, we adopt the idea by using Q-LoRA to fine-tune an LLM as a sign language translator. We adopted the Qwen2 LLM model as our translator. To fine-tune Qwen2, we need to set the LLM using the scenario as a "Sign language translator" and design prompts to guide the model. In the prompts, we provide the top 3 gloss sequences mentioned in \ref{seq:slr} and all the above text related to the current sign language sequence, and ask the LLM to summarize the top 3 glosses and guess the correct words to use by checking previous texts. We also provide some summarized task examples to help the LLM understand translation procedures. We use the previous top 3 gloss outputs as input and use the designed prompt along with the above text as auxiliary input, jointly fine-tuning the LLM model. We optimize the model using the cross-entropy loss function:
\begin{equation} \label{eq:llmcrossentropyloss}
\mathcal{L}_\textit{llm} = -\sum_{i} \hat{Y}_t(i) \log(Y_t(i)), i \in N_\textit{tok}.
\end{equation}
$\hat{Y}_t$ is the ground truth textual output, and $Y_t$ is the predicted textual output. $N_{tok}$ is the number of classes in the tokenizer.

\begin{figure*}[t]
	\centering
	\includegraphics[width=\linewidth]{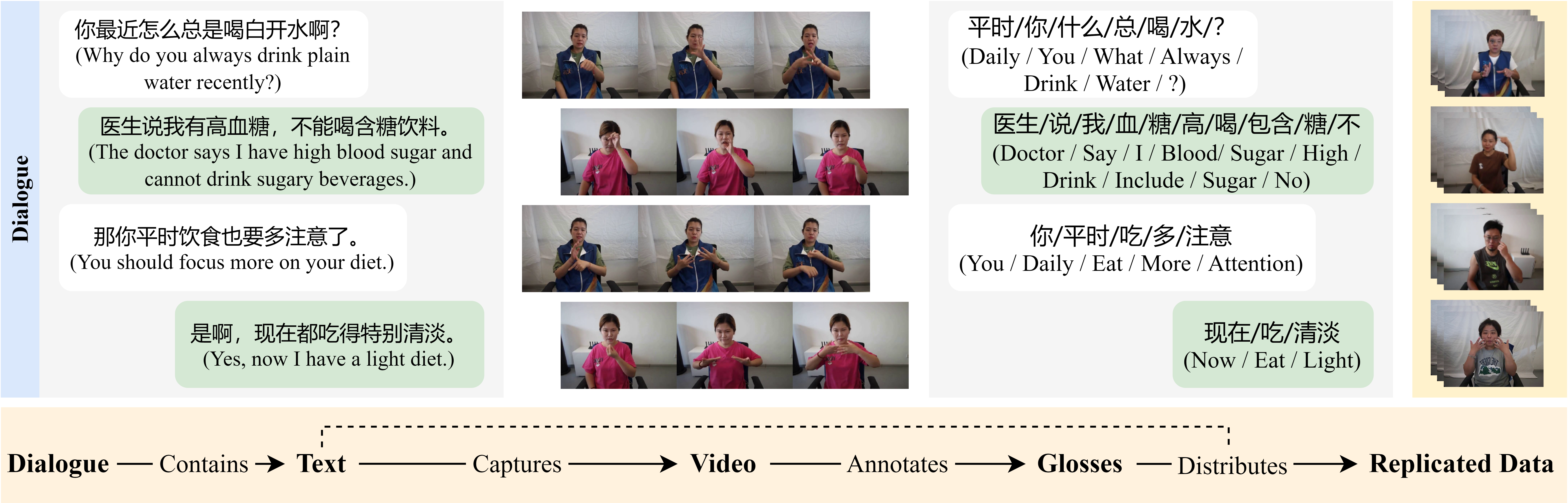}
        \vspace{-10pt}
	\caption{\textbf{SCOPE dataset collection pipeline.} Given dialogue texts, experienced signers produce corresponding sign videos along with self-annotated glosses. For each video, other signers replicate data based on the glosses and the text.} 
	\label{img:dataset}
\end{figure*}
\vspace{20pt}
\begin{table*}[t]
\centering
\begin{tabular}{l|l|l|l|l|l|l|l|l|l} 
\hline
Dataset &Language & Videos & Duration(h) & Signers & Vocab & Gloss & Text & Dialogue &Source  \\
\hline
PHOENIX-2014  &DGS & 6,841 & 8.9 & 9 & 1k  (German)   & \checkmark &  $\times$  & $\times$ &TV   \\
PHOENIX-2014T &DGS & 8,257  & 11 & 9 & 3k  (German)   & \checkmark &  \checkmark  & $\times$ &TV  \\
CSL-Daily  &CSL & 20,654 & 22.8 & 10 & 2k  (Chinese)   & \checkmark &  \checkmark  & $\times$ &Lab \\
How2Sign &ASL & 35,191  & 79 & 11 & 16k(English)   & \checkmark &  \checkmark  & $\times$  &Lab  \\
\hline
Ours &CSL & 59,231  & 72.4 & 12 & 5k  (Chinese)   & \checkmark &  \checkmark  & \checkmark  &Lab \\
\hline
\end{tabular}
\vspace{1mm}
\caption{\textbf{Dataset comparisons.} Key statistics of widely used sign language datasets for comparison. Our dataset is currently the largest dataset in CSL (Chinese Sign Language) that contains dialogue context information.
}
\label{tab:dataset}
\vspace{-7pt}
\end{table*}

\subsection{Data Processing}

\myparagraph{Iris Normalization.}
To fetch keypoint sequences, we utilized DWPose \cite{yang2023effective} to obtain whole-body keypoints (COCO-WholeBody \cite{jin2020whole}) from sign language videos. Each keyframe contains 133 keypoints $J = \{J_{1,i},..., J_{T,i}|i=1...133\}$. However, such keypoints are often influenced by the input video resolution and the distance between the person and the camera. A scaling process is needed to mitigate the impact of input distortions on motion. Inspired by \cite{lugaresi2019mediapipe}, we choose the length of the lower eyelid in humans as the golden standard, comparing the eyelid length differences to get the scale factor and scale motions to the standard size:
\begin{equation} \label{eq:scale}
J^\textit{scaled}_{t} = \frac{(J_{x},J_{y})}{|(J_{x_\textit{63}}-J_{x_\textit{64}})|},
\end{equation}
Where $J^\textit{Scaled}_{t}$ are scaled joints under frame $t$, $(J_{x},J_{y})$ are 2D coordinates of joints. $J_{x_\textit{63}}-J_{x_\textit{64}}$ is eyelid length; 63 and 64 are indexes of left and right wings of the eyelid in COCO-WholeBody. 

\myparagraph{Data Centralizing.}
After that, we followed \cite{jiao2023cosign} by selecting 77 keypoints and dividing them into 5 groups, then applied group-specific centralization to decouple multi-grained motion information in skeleton data:
\begin{equation} \label{eq:centeralize}
J_{t,k} = J_{t,k} - J_{t,r_{g}}, k \in G,
\end{equation}
where $J_{t,k}$ denotes joints under the $t$ frame, group $k$, $G$ are 5 groups, and $r_{g}$ is the root keypoint of group $g$. 

\myparagraph{Data Standardizing.}
Finally, we standardize all input motions to make their distribution more closely conform to a standard distribution, which eliminates the difficulties that motion corner cases bring to training:
\begin{equation} \label{eq:standarlize}
J_{i}^\textit{std} = \mathcal{N}(J_{i} - \frac{\sum\limits^{n}\sum\limits^{t}J_{i}}{N \times T},\frac{I}{J_{i}^\textit{std}}), i=1...77 ,
\end{equation}
where $J_{i}$ denotes the $i-th$ joints, and $J^\textit{std}$ is the standard deviation of joint $i$.

\section{SCOPE dataset}

A sign language dataset with contextual information is essential to fully leverage the power of context in implementing our context-aware approaches. We propose SCOPE, a large-scale Chinese sign language dataset that includes contextual dialogue information. Our data information and dataset comparison can be found in Tab.\ref{tab:dataset}.

\myparagraph{Data Collection.}
Our dataset primarily focuses on daily conversations within the Deaf community, as well as dialogues involving specialized terminology in more professional settings\cite{bragg2019sign}. Our dataset includes daily subjects such as school experiences, shopping, and social interactions. Glosses encompass specific products and brands, titles of audiovisual works, and other relevant terms. For more details, please refer to the supplementary material.

Data collection is carried out by a team whose primary members are several professional Deaf signers and three sign language linguistics experts. The team also includes a diverse group of Deaf individuals across various ages, genders, occupations, and educational backgrounds to capture diverse signing styles. To ensure a natural dialogue environment, each sentence was derived from conversations recorded in real situations.

Fig.\ref{img:dataset} illustrates our data collection pipeline. Professional Deaf signers receive reference sentences and record corresponding sign language videos. Capable of self-annotating recorded motion into glosses, they produce gloss annotations that are distributed to other signers. With sentences and glosses as references, other signers replicate data with diverse signing habits and styles. We ensure that four different signers record videos for each piece of text at a resolution of 640x480 and a frame rate of 30 frames per second.

\myparagraph{Annotation Cleaning and Validation.}
Self-annotated glosses still suffer from inconsistencies across different annotators. A same sign is sometimes annotated with synonyms, while a sequence of signs may get interpreted into phrases or separated words. To mitigate such issues, we follow CSL-Daily\cite{zhou2021improving} to apply a multi-round data cleaning process with our SCOPE SLR model. 

Particularly, we compute Minimum Edit Distance (MED) between predicted glosses and ground truth from annotation. The results enable us to identify patterns of synonyms, word division and word combination. Our sign language linguistics experts then examine frequently confused patterns and correct misannotated data. We iterate such a process to reduce our gloss vocabulary size from over 7k to 5k, significantly improving the dataset's quality.
 
\section{Experiments}

\begin{table*}[t]

    \centering
    \resizebox{0.8\linewidth}{!}{
    \begin{tabular}{l|cc|cc|cc|cc}
    \toprule
        \multirow{2}{*}{Method}  & \multicolumn{2}{c|}{Phoenix-2014} & \multicolumn{2}{c|}{Phoenix-2014T} & \multicolumn{2}{c|}{CSL-Daily} & \multicolumn{2}{c}{SCOPE} \\
        \cmidrule(r){2-3} \cmidrule(r){4-5} \cmidrule(r){6-7} \cmidrule(r){8-9}
        & Dev & Test &  Dev & Test &   Dev & Test &   Dev & Test \\
        \midrule
        SEN-CSLR\cite{hu2023self} & 19.5 & 20.9 & 19.3 & 20.7 &31.1&30.7 & 40.2 & 41.1 \\
        TwoStream-SLR\cite{chen2022two} & \textcolor{red}{\textbf{18.4}} & \textcolor{red}{\textbf{18.8}} & \textcolor{red}{\textbf{17.7}} & \textcolor{blue}{\textbf{19.3}} &\textcolor{blue}{\textbf{25.4}}&\textcolor{blue}{\textbf{25.3}} & 40.8 & 40.5 \\
        CorrNet\cite{hu2023continuous} & 18.9 & 19.7 &18.9&20.5 & 30.6 & 30.1   & 33.5 & 33.8 \\
        \midrule
        Ours-SLR$^*$ w/o Context &\textcolor{blue}{\textbf{18.8}}&\textcolor{blue}{\textbf{19.2}}&\textcolor{blue}{\textbf{17.8}}&\textcolor{red}{\textbf{19.0}}&\textcolor{red}{\textbf{22.7}}&\textcolor{red}{\textbf{23.1}} & \textcolor{blue}{\textbf{30.2}} & \textcolor{blue}{\textbf{30.7}}\\
        Ours-SLR$^*$ &-&-&-&-&-&- & \textcolor{red}{\textbf{28.0}} & \textcolor{red}{\textbf{27.4}}\\
    \bottomrule
    \end{tabular}}
    \vspace{-3pt}
    \caption{\textbf{Quantitative evaluation} of \textbf{Sign Language Recognition (SLR)} task. WER is adopted as the evaluation metric. We train other methods and ours on our SCOPE dataset. Also, our model without context input is evaluated on other popular datasets. The \textcolor{red}{red} and \textcolor{blue}{blue} entries indicate the best and the second best results.}
    \label{tab:baseline_slr}
    \vspace{5pt}
\end{table*}

\begin{table*}[t]
\centering
\resizebox{\linewidth}{!}{
\small
\begin{tabular}{llcccccccccc}
\toprule
\multirow{2}{*}{Dataset} & \multirow{2}{*}{Method} & \multicolumn{5}{c}{Dev} & \multicolumn{5}{c}{Test} \\
\cmidrule(r){3-7} \cmidrule(r){8-12}
    & & R$\uparrow$ & B1$\uparrow$ & B2$\uparrow$ & B3$\uparrow$ & B4$\uparrow$ & R$\uparrow$ & B1$\uparrow$ & B2$\uparrow$ & B3$\uparrow$ & B4$\uparrow$ \\
\midrule

\multirow{5}{*}{P-2014T} 
& MMTLB-S2T\cite{chen2022simple}  & 53.10 & 53.95 & 41.12 & 33.14 & 27.61 & 52.65 & 53.97 & 41.75 & 33.84 & 28.39 \\
& TwoStream-S2T\cite{chen2022two}  & 54.08 & 54.32 & 41.99 & 34.15 &28.66 & 53.48 & \textcolor{blue}{\textbf{54.90}} & 42.43 & 34.46 & 28.95\\
& CV-SLT\cite{zhao2024conditional}  & \textcolor{blue}{\textbf{54.43}} & \textcolor{blue}{\textbf{55.09}} & \textcolor{blue}{\textbf{42.60}} & \textcolor{blue}{\textbf{34.63}} & \textcolor{blue}{\textbf{29.10}} & \textcolor{blue}{\textbf{54.33}} & 54.88 & \textcolor{blue}{\textbf{42.68}} & \textcolor{blue}{\textbf{34.79}} & \textcolor{blue}{\textbf{29.27}} \\

& Ours$^*$ w/o Context  & $\textcolor{red}{\textbf{67.09}}$ & \textcolor{red}{\textbf{61.80}} & \textcolor{red}{\textbf{49.09}} & \textcolor{red}{\textbf{39.53}} & \textcolor{red}{\textbf{32.83}} & \textcolor{red}{\textbf{60.06}} & \textcolor{red}{\textbf{61.74}} & \textcolor{red}{\textbf{49.22}} & \textcolor{red}{\textbf{39.61}} & \textcolor{red}{\textbf{32.84}} \\
\midrule

\multirow{5}{*}{CSL-Daily} 

& MMTLB-S2T\cite{chen2022simple}  & 53.38 & 53.81 & 40.84 & 31.29 & 24.42 & 53.25 & 53.31 & 40.41 & 30.87 & 23.92 \\
& TwoStream-S2T\cite{chen2022two}  & 55.10 & 55.21 & 42.31 & 32.71 &25.76 & 55.72 & 55.44 & 42.59 & 32.87 & 25.59\\
& CV-SLT\cite{zhao2024conditional}  & \textcolor{blue}{\textbf{56.36}} & \textcolor{blue}{\textbf{58.05}} & \textcolor{blue}{\textbf{44.73}} & \textcolor{blue}{\textbf{35.14}} & \textcolor{blue}{\textbf{28.24}} & \textcolor{blue}{\textbf{57.06}} & \textcolor{blue}{\textbf{58.29}} & \textcolor{blue}{\textbf{45.15}} & \textcolor{blue}{\textbf{35.77}} & \textcolor{blue}{\textbf{28.94}} \\
& Ours$^*$ w/o Context  & \textcolor{red}{\textbf{60.18}} & \textcolor{red}{\textbf{60.37}} & \textcolor{red}{\textbf{47.21}} & \textcolor{red}{\textbf{37.36}} & \textcolor{red}{\textbf{31.74}} & \textcolor{red}{\textbf{60.68}} & \textcolor{red}{\textbf{60.48}} & \textcolor{red}{\textbf{49.61}} & \textcolor{red}{\textbf{40.01}} & \textcolor{red}{\textbf{32.08}}  \\
\midrule

\multirow{5}{*}{SCOPE} 
& MMTLB-S2T\cite{chen2022simple}   & 63.25 & 60.72 & 50.33 & 40.39 & 31.61 & 64.30 & 61.69 & 51.75 & 41.98 & 33.56 \\
& TwoStream-S2T\cite{chen2022two}  & 63.40 & 60.87 & 50.48 & 40.74 &31.65 & 64.30 & 61.78 & 51.86 & 42.17 & 33.50\\
& CV-SLT\cite{zhao2024conditional}  & 65.71 & 63.16 & 52.00 & \textcolor{blue}{\textbf{43.69}} & \textcolor{blue}{\textbf{37.10}} & 66.06 & 62.69 & 52.12 & \textcolor{blue}{\textbf{44.14}} & \textcolor{blue}{\textbf{37.82}} \\
& Ours$^*$ w/o Context  & \textcolor{blue}{\textbf{69.34}} & \textcolor{blue}{\textbf{64.31}} & \textcolor{blue}{\textbf{53.15}} & 43.57 & 34.83 & \textcolor{blue}{\textbf{69.46}} & \textcolor{blue}{\textbf{64.62}} & \textcolor{blue}{\textbf{53.64}} & 44.13 & 35.80 \\
& Ours$^*$  & \textcolor{red}{\textbf{69.78}} & \textcolor{red}{\textbf{65.68}} & \textcolor{red}{\textbf{55.06}} & \textcolor{red}{\textbf{46.18}} & \textcolor{red}{\textbf{38.09}} & \textcolor{red}{\textbf{70.14}} & \textcolor{red}{\textbf{65.85}} & \textcolor{red}{\textbf{55.42}} & \textcolor{red}{\textbf{46.56}} & \textcolor{red}{\textbf{38.59}} \\

\bottomrule
\end{tabular}
}
\vspace{1mm}
\caption{\textbf{Quantitative evaluation} of \textbf{Sign Language Translation (SLT)} task. (R: ROUGE, B: BLEU) We train other methods on our dataset, our method on all three datasets. For non-context-based data, we train our method without context. The \textcolor{red}{red} and \textcolor{blue}{blue} entries indicate the best and the second-best results.}
\label{tab:sign_slt}
\end{table*}


\subsection{Experimental Setup}
\myparagraph{Datasets and Evaluation Metrics.}
For the SLR task, we evaluate our proposed method on PHOENIX14, PHOENIX14-T, CSL-Daily, and SCOPE dataset; the latter three datasets are also utilized in experiments on SLT task. For the number of samples and other details of datasets, please refer to supplementary materials.
Train/dev/test splits of the existing datasets are maintained. For our SCOPE dataset, we follow \cite{zhang2024both2hands} to use widely adopted split ratios to randomly split our dataset by 80\%, 5\% and 15\% into train, dev, and test sets, carefully ensuring that no same sentence appears in different sets and any sentence in the dev set or test set does not appear in context dialogues of the training set.


	


Following previous works \cite{chen2022two}, we adopt the Word Error Rate (WER), which measures the percentage of incorrect words in recognized text, for SLR, and BLEU \cite{papineni2002bleu} and ROUGE-L \cite{lin2004rouge}, which assess the quality of translations based on n-gram overlap and longest common subsequences, for SLT as evaluation metrics. Lower WER indicates more accurate recognition results, while higher BLEU and ROUGE-L signify better translations.

\myparagraph{Implementation Details.}
We obtain sentence embeddings by OpenAI's text-embedding-ada-002 \cite{neelakantan2022text} model. Body 2D keypoints are collected from videos using DWPose \cite{yang2023effective}. Our motion feature extractor block consists of a 2-layer MLP with a temporal Conv1D layer. 
The embedding alignment encoder and gloss encoder are both 8-head transformer encoders with 2 and 4 layers, respectively, with hidden size 1568 and feed-forward size 3136. 
We adopt the AdamW optimizer and use cosine annealing schedules, with 20 epochs focusing on alignment embedding, and 60 epochs for gloss encoder training while keeping the previous module frozen. 
When training without the context module, we do not provide context information by filling context embeddings with zeros and providing empty context input for LLM. 
All experiments are executed on 8 NVIDIA A800 GPUs. More implementation details are provided in the supplementary materials.

\subsection{Comparison with State-of-the-art Methods}
\myparagraph{Sign Language Recognition (SLR)}
We evaluate our approach by comparing results on multiple datasets with recent methods SEN-CSLR\cite{hu2023self}, TwoStream-SLR\cite{chen2022two} and CorrNet\cite{hu2023continuous}.  

On our SCOPE dataset, we evaluate their performance by training their open-sourced framework. We perform preprocessing to match the input specifications of each method and train their models adhering as closely as possible to their proposed training setups. 

As shown in Tab.\ref{tab:baseline_slr}, our context-free SCOPE outperforms other SLR methods in WER by \textbf{2.7}\%/\textbf{2.2}\% on CSL-Daily dev/test sets and \textbf{3.3}\%/\textbf{3.1}\% on SCOPE dataset, respectively.
Moreover, adding context information further improves our model's recognition accuracy by \textbf{2.2}\%/\textbf{3.3}\% WER, revealing that contextual understanding effectively assists gloss recognition. 

\myparagraph{Sign Language Translation (SLT)}
On the SLT task, we compare our approach with state-of-the-art gloss-supervised and gloss-free methods. Similarly, we stick to their respective training configurations when training their models on SCOPE dataset.
Results in Tab.\ref{tab:sign_slt} show that our approach outperforms previous methods by +3.73/+3.57 BLEU and +3.50/+3.14 BLEU in Phoenix-2014T and CSL-Daily dev/test sets. Additionally, our full approach on SCOPE dataset brings another +3.26/+2.79 BLEU improvement, which we attribute mainly to context-aware LLM fine-tuning. Notably, when comparing across datasets, SCOPE dataset generally yields better performance for any fixed method. We primarily attribute this result to our robust data annotation and cleaning process.
\subsection{Ablation Studies}
We conduct ablation experiments for both SLR and SLT tasks to validate the contributions of each component. 
The comparison between our full and context-free SCOPE model also suffices as an ablation study to demonstrate the significance of context information, both in recognition and in LLM fine-tuning.
When the embedding alignment encoder is removed, the context sentence embeddings are concatenated to motion features directly, and $\mathcal{L}_\textit{emb}$ no longer serves as a supervision term. The performance of this model declines by 4.4\%WER and 16.01 BLEU, and we note that it takes significantly longer for this model to converge. Thus, we deduce that the model encounters difficulties in aligning motion features with LLM context embeddings and ultimately behaves poorly. The removal of $\mathcal{L}_\textit{MWER}$ directly causes more word errors, thus deteriorating the translation results. 
The distribution of raw keypoints is severely biased without our Iris Normalization process, rendering the model overfit to extreme cases and unsuitable for real-time practical use with different aspect ratios and camera resolutions.


\subsection{Real-time Application and User Studies}
Authentic feedback from the Deaf community is the gold standard for practical use. We have developed a real-time SLT application to assist Deaf individuals in accessing dental care. Details are provided in the supplementary materials.  
Authentic feedback from the Deaf community is the gold standard for practical use. We have developed a real-time SLT application to assist Deaf individuals. Details are provided in the supplementary materials.  
We conducted a survey on their user experience, and questions concerning random SLR or SLT results. We have collected 40 responses, rating our application user experience as 4.15 / 5 on average, accuracy of SLR results as 3.96 / 5, and SLT results as 3.98 / 5. These ratings indicate a positive response from the Deaf community, providing strong evidence of our research's effectiveness.

\begin{table}[t]
\vspace{3mm}
\centering
\resizebox{\linewidth}{!}{
\small
\begin{tabular}{lc|cc}
\toprule
\multirow{2}{*}{Ablation Study} & SLR & \multicolumn{2}{c}{SLT}  \\
\cmidrule(r){2-4}
    & WER$\downarrow$ & R$\uparrow$ & B4$\uparrow$\\
\midrule
Full SCOPE$^*$  & \textbf{27.4} & \textbf{70.14} & \textbf{38.59}  \\
w/o Context  & 30.7 & 69.46 & 35.80   \\
w/o Embedding Encoder  & 31.8 & 51.77 & 22.58 \\
w/o $\mathcal{L}_\textit{MWER}$  & 37.6 & 48.64 & 15.55 \\
w/o Iris Normalization  & 35.8 & 51.51 & 21.76 \\

\bottomrule
\end{tabular}
}
\vspace{0mm}
\caption{\textbf{Ablation studies of our contextual design and data processing algorithm.}}
\vspace{-5mm}
\label{tab:sign_abl}
\end{table}
 
\section{Conclusion}
We present the SCOPE dataset, the first dialogue-based Chinese Sign Language dataset featuring both gloss and text annotations. This dataset encompasses 72.4 hours of sign language videos collected from professional Deaf groups, complemented by 59,231 text annotations. Building on this dataset, we introduce the SCOPE framework, a robust pipeline specifically designed to address Sign Language Recognition (SLR) and Sign Language Translation (SLT) tasks with rich contextual information. Our comprehensive evaluations demonstrate the effectiveness of our methods and the significant improvements enabled by our dataset for the sign language community. We believe that SCOPE will catalyze future research in context-based sign language processing.
\vspace{3mm}

\bibliography{arxiv}

\section{Appendix}

\subsection{SCOPE Dataset details}

\myparagraph{Data Collection Details}
The SCOPE dataset was meticulously curated to ensure high-quality and diverse sign language data. The data collection process involved several steps and considerations:

\begin{itemize}
    \item \textbf{Participants}: Our participants included both professional Deaf sign language teachers and non-professional Deaf individuals. The professional signers comprised three sign language linguistics experts and several experienced Deaf signers. Non-professional signers were selected to represent a variety of ages, genders, occupations, and educational backgrounds, capturing a wide range of signing habits.
    
    \item \textbf{Location}: Data collection sessions were conducted in a controlled environment designed to mimic real-life scenarios. This setting was equipped with high-resolution cameras and appropriate lighting to ensure clarity and accuracy in the captured sign language videos.
    
    \item \textbf{Collection Process}: 
    \begin{enumerate}
        \item Professional signers were provided with reference sentences in natural language and asked to perform the corresponding sign language. These sessions were recorded, and the signers annotated the videos to create initial gloss annotations.
        \item The annotated videos were reviewed by our sign language linguistics experts to ensure accuracy and consistency. Any discrepancies identified during this phase were corrected.
        \item Non-professional signers were given the gloss annotations and reference sentences to replicate the sign language videos. This step ensured the inclusion of diverse signing styles and habits.
        \item Each reference sentence was performed by four different signers, resulting in multiple video samples per sentence to enhance the richness of the dataset.
    \end{enumerate}
    
    \item \textbf{Data Cleaning}: Our data cleaning process consists of two main steps.
    \begin{enumerate}
        \item Three sign language linguistics experts reviewed all gloss annotations. Utilizing their expertise, they identified equivalent gloss combinations, including one-to-one, one-to-many, and many-to-one relationships. One-to-one relationships were generally synonymous, and we consolidated these synonyms into a single gloss. One-to-many and many-to-one relationships often pertained to phrases, and we determined the use of phrases based on the frequency of gloss occurrences.
        \item We developed a script to automate the identification and correction of inconsistencies in sign language annotations. This script analyzed discrepancies between the test set and the ground truth, identifying common error types derived from the calculation of Word Error Rate (WER), such as "C-S-I-C", "C-I-S-C", "C-D-S-C", and "C-S-D-C". In these acronyms, each letter represents a specific error type: Correct (C), Insertion (I), Substitution (S), and Deletion (D). Linguistics experts then examined and corrected frequently confused patterns. Additionally, the script maintained a record of previously processed annotation pairs to avoid redundancy, thereby improving the efficiency and accuracy of the data cleaning process. Through this iterative process, we successfully reduced the vocabulary size from 7,000 to 5,000, significantly enhancing the quality of the dataset. 
    \end{enumerate}
    
    \item \textbf{Database Management}: All collected data, including raw videos, annotations, and metadata, were stored in a structured database. This database facilitated easy access and management of the dataset for further processing and analysis.
\end{itemize}

\begin{table}[t]
\centering
\renewcommand\arraystretch{1.1}
\resizebox{0.85\linewidth}{!}{
\begin{tabular}{lll}
\hline
Scene Statistics &Clip Num & Time  \\ \hline
Medical treatment       &1,3265      & 18.24h \\
Workplace     &12,597        & 17.91h \\
Entertainment  &10,391      & 15.10h \\ 
Family        &10,775 & 10.68h \\
Education  &10,740      & 8.46h \\ 
Shopping  &1,457      & 2.01h \\ \hline
Total      &59,231             & 72.40h \\ \hline
\end{tabular}
}
\vspace{-3mm}
\caption{\textbf{Dataset Statistics.} Scene Statistics classified scenario appeared in the dataset, and calculate their total length. }
\label{tab:DataStatistics}
\end{table}

\begin{figure*}[h]
    \centering
    \includegraphics[width=0.9\textwidth]{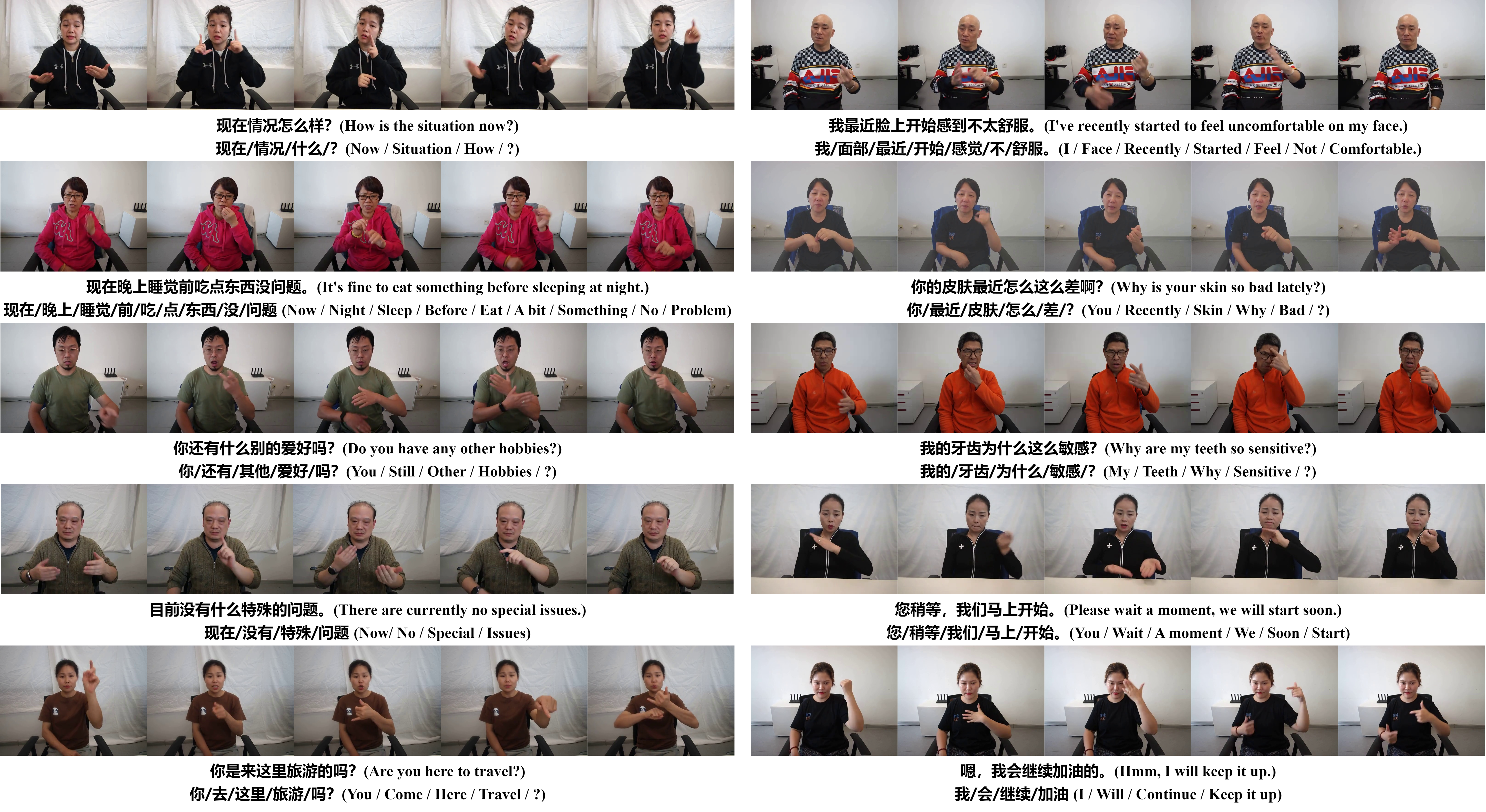}
    \caption{\textbf{SCOPE gallery.} We sampled different scenarios and show case the dataset sign videos and annotations.}
    \label{fig:gallery}
\end{figure*}

\begin{figure}[h]
    \centering
    \includegraphics[width=\linewidth]{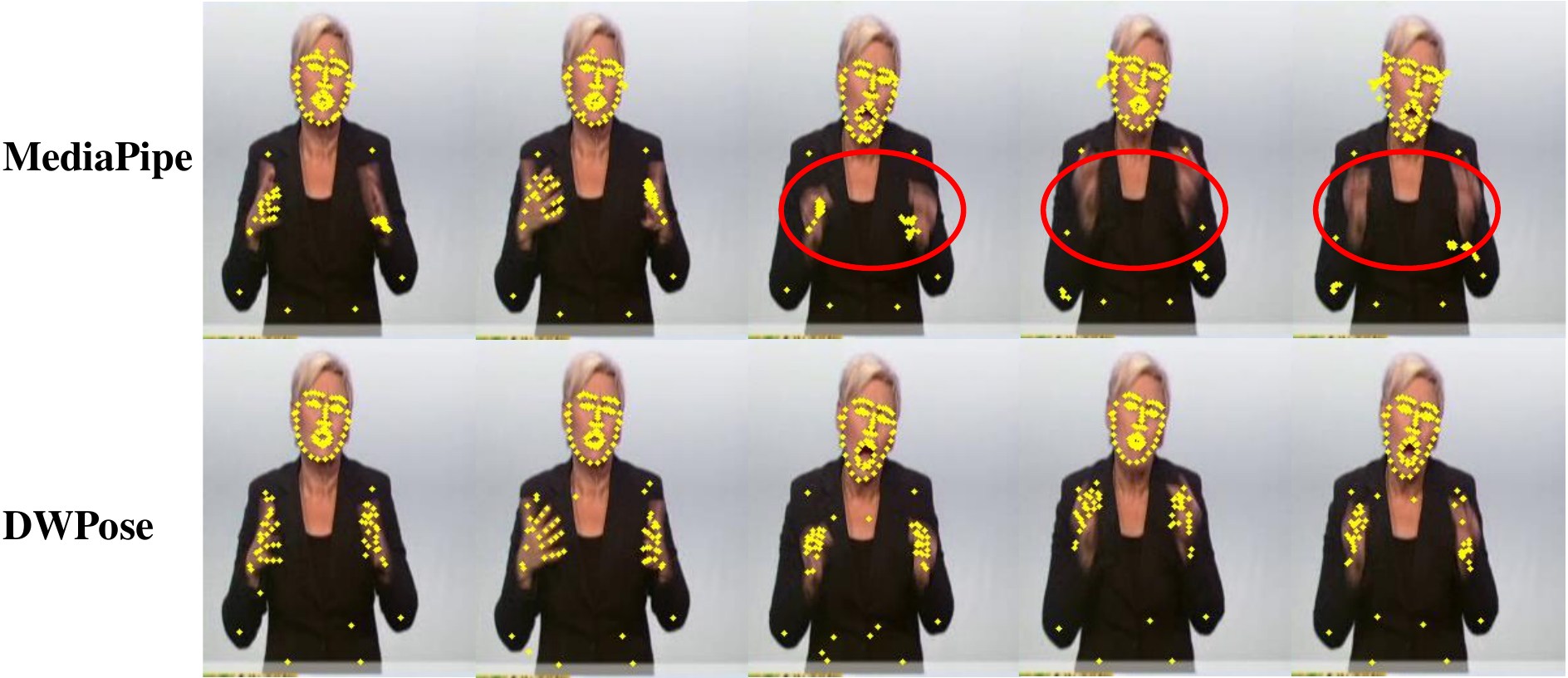}
    \caption{\textbf{Comparison of 2D keypoints} identified by MediaPipe (up) and DWPose (down) from the Phoenix 2014 dataset. DWPose provides more accurate and detailed keypoints.}
    \label{fig:pose_comparison}
\end{figure}

\myparagraph{Data Statistics}
The SCOPE dataset covers a wide range of scenarios and dialogue contexts, as detailed in Tab.\ref{tab:DataStatistics}. It includes professional settings such as medical treatment, work, and education, as well as daily situations like entertainment, family communication, and shopping. The dataset contains 33,154 clips under 5 seconds and 26,077 clips over 5 seconds, with an average length of 5.05 seconds. The average sentence length is 11.84 characters, highlighting the dataset's richness. Fig.\ref{fig:gallery} presents visual examples from the SCOPE dataset, showcasing its diversity.

\subsection{Experiment Details}
In this section, we provide an in-depth discussion of the experiment's details, including input data quality, preparation specifics, and additional evaluation comparisons.

\myparagraph{Input Details}
We employed the state-of-the-art Pose Estimation method, DWPose \cite{yang2023effective}, to accurately identify keypoint positions for each action frame. Fig.\ref{fig:pose_comparison} compares the 2D keypoints identified by MediaPipe \cite{lugaresi2019mediapipe} from the POENIX-2014 dataset with those identified by DWPose. The keypoints from DWPose demonstrate significantly higher quality than those from previous methods.

\myparagraph{Setup Details}
The experimental setup was meticulously designed to ensure the robustness and reproducibility of our results. We begin by introducing the datasets used in the experiment:

\begin{itemize} 
\setlength\itemsep{0em}
	\item \textbf{PHOENIX14} is a German sign language dataset focused on weather forecasts, containing 5,672 training samples, 540 development samples, and 629 test samples. 

	\item \textbf{PHOENIX14-T} extends PHOENIX14 with both gloss and translation annotations, including 7,096 training samples, 519 development samples, and 642 test samples.
	
	\item \textbf{CSL-Daily} is a Chinese sign language dataset with gloss and translation annotations, comprising 18,401 training samples, 1,077 development samples, and 1,176 test samples.

    \item \textbf{SCOPE}, as previously mentioned, is a Chinese sign language dataset featuring gloss annotations, translation annotations, and dialogue context information, with a total of 59,231 samples across training, development, and test sets.
\end{itemize} 

Experiments were conducted on a cluster of 8 NVIDIA A800 GPUs. Each GPU was assigned a subset of the dataset to facilitate parallel processing and efficient resource utilization. All sign language videos were preprocessed to extract 2D keypoints using DWPose, which were then normalized using our Iris Normalization technique to account for variations in video resolution and camera distance. Training details are provided below:

\begin{itemize}
\item \textbf{SCOPE}: As mentioned in the previous section.
\item \textbf{SCOPE without context}: Most settings are similar to the full pipeline, but we use random noise instead of context embedding for the embedding alignment encoder and mask their attention. For the LLM, the prompt was modified to "Summarizing the sentence using top 3 glosses only."
\item \textbf{SEN-CSLR, CorrNet, and CV-SLT}: We pretrained the models on our dataset following their detailed instructions.
\item \textbf{Two-Stream SLR}: We first applied single-stream pretraining on our video and keypoint data modalities, followed by two-stream joint training to predict gloss results.
\item \textbf{MMTLB and Two-Stream SLT}: For MMTLB and the Two-Stream Network, we pretrained the sign-to-gloss module and extracted visual features before performing joint training for the final translation process.
\end{itemize}

\begin{table*}[h]
    \centering
    \resizebox{\textwidth}{!}{
    \begin{tabular}{l|cc|cc|cc|cc}
    \toprule
        \multirow{2}{*}{Method}  & \multicolumn{2}{c|}{Phoenix-2014} & \multicolumn{2}{c|}{Phoenix-2014T} & \multicolumn{2}{c|}{CSL-Daily} & \multicolumn{2}{c}{SCOPE} \\
        \cmidrule(r){2-3} \cmidrule(r){4-5} \cmidrule(r){6-7} \cmidrule(r){8-9}
        & Dev & Test &  Dev & Test &   Dev & Test &   Dev & Test \\
        \midrule
        
        SubUNets \cite{cihan2017subunets} & 40.8 & 40.7 & - & - & 41.4 & 41.0 & - & -\\
        LS-HAN \cite{huang2018video} & - & - & - & - & 39.0 & 39.4 & - & -\\
        IAN \cite{pu2019iterative} & 37.1 & 36.7 & - & - & - & - & - & -\\
        ReSign  \cite{koller2017re} & 27.1 & 26.8 & - & - & - & - & - & -\\
        CNN-LSTM-HMMs (Multi-Stream) \cite{koller2019weakly} & 26.0 & 26.0 & 22.1 & 24.1 & - & - & - & -\\
        SFL \cite{niu2020stochastic} & 24.9 & 25.3 & 25.1 & 26.1 & - & - & - & -\\
        DNF (RGB) \cite{cui2019deep} & 23.8 & 24.4 & - & - & 32.8 & 32.4 & - & -\\
        FCN \cite{cheng2020fully} & 23.7 & 23.9 & 23.3 & 25.1 & 33.2 & 33.5 & - & -\\
        DNF (RGB+Flow) \cite{cui2019deep} & 23.1 & 22.9 & - & - & - & - & - & -\\
        Joint-SLRT \cite{camgoz2020sign} & - & - & 24.6 & 24.5 & 33.1 & 32.0 & - & -\\
        VAC \cite{min2021visual} & 21.2 & 22.3 & - & - & - & - & - & -\\
        LCSA \cite{zuo2022local} & 21.4 & 21.9 & - & - & - & - & - & -\\
        CMA \cite{pu2020boosting} & 21.3 & 21.9 & - & - & - & - & - & -\\
        SignBT \cite{zhou2021improving} & - & - & 22.7 & 23.9 & 33.2 & 33.2 & - & -\\
        MMTLB \cite{chen2022simple} & - & - & 21.9 & 22.5 & - & - & - & -\\
        SMKD \cite{hao2021self} & 20.8 & 21.0 & 20.8 & 22.4 & - & - & - & -\\
        STMC-R (RGB+Pose) \cite{zhou2021spatial} & 21.1 & 20.7 & 19.6 & 21.0 & - & - & - & -\\
        $C^{2}$SLR (RGB+Pose) \cite{zuo2022c2slr} & 20.5 & 20.4 & 20.2 & 20.4 & - & - & - & -\\
        
        \midrule
        SEN-CSLR \cite{hu2023self} & 19.5 & 20.9 & 19.3 & 20.7 & 31.1 & 30.7 & 40.2 & 41.1 \\
        TwoStream-SLR \cite{chen2022two} & 18.4 & 18.8 & \textcolor{blue}{\textbf{17.7}} & 19.3 & \textcolor{blue}{\textbf{25.4}} & 25.3 & 40.8 & 40.5 \\
        SlowFastSign \cite{ahn2024slowfast} & \textcolor{red}{\textbf{18.0}} & \textcolor{red}{\textbf{18.2}} & \textcolor{red}{\textbf{17.6}} & \textcolor{red}{\textbf{18.7}} & 25.4 & \textcolor{blue}{\textbf{24.8}} & 35.2 & 35.6 \\
        CorrNet \cite{hu2023continuous} & 18.9 & 19.7 & 18.9 & 20.5 & 30.6 & 30.1 & 33.5 & 33.8 \\
        \midrule
        Ours-SLR$^*$ w/o Context & \textcolor{blue}{\textbf{18.8}} & \textcolor{blue}{\textbf{19.2}} & 17.8 & \textcolor{blue}{\textbf{19.0}} & \textcolor{red}{\textbf{22.7}} & \textcolor{red}{\textbf{23.1}} & \textcolor{blue}{\textbf{30.2}} & \textcolor{blue}{\textbf{30.7}} \\
        Ours-SLR$^*$ & - & - & - & - & - & - & \textcolor{red}{\textbf{28.0}} & \textcolor{red}{\textbf{27.4}} \\
    \bottomrule
    \end{tabular}}
    \caption{\textbf{Quantitative evaluation} of various Sign Language Recognition (SLR) tasks. Word Error Rate (WER) is used as the evaluation metric. We trained both our method and other approaches on the SCOPE dataset. Additionally, our model without context input was evaluated on other popular datasets. The \textcolor{red}{red} entries indicate the best results, while the \textcolor{blue}{blue} entries represent the second-best results.}
    \label{tab:SLR_comparison_more}
\end{table*}

\begin{figure}[h]
    \centering
    \includegraphics[width=1.0\linewidth]
    {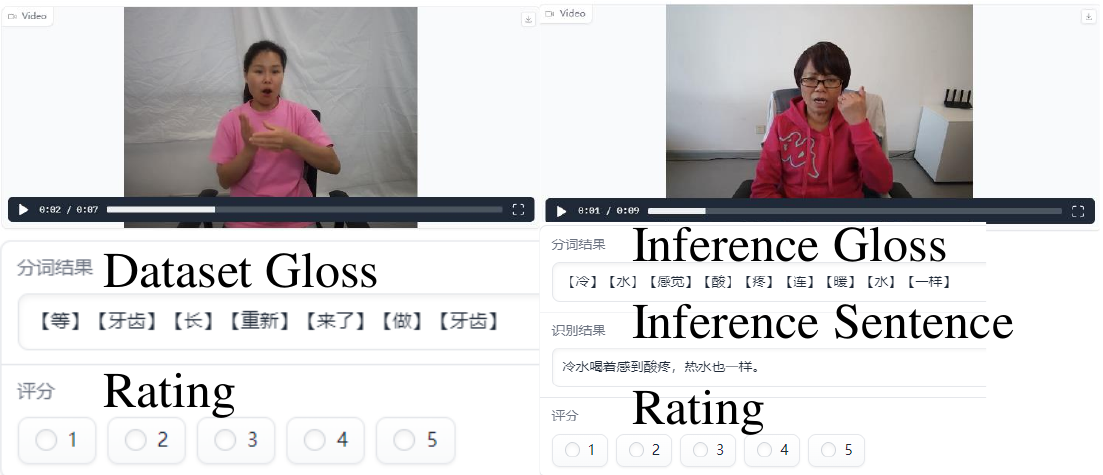}
    \caption{\textbf{User study on dataset and inference quality.} In our user interface, participants rank the quality of the dataset gloss (upper right) and the inference results (lower right) in relation to the sign video (upper left and lower left). A higher rank indicates better quality.}
    \label{fig:user_study}
\end{figure}

\myparagraph{More Evaluation}
To demonstrate the robustness of our pipeline, we conducted extensive evaluations comparing our SCOPE framework with several existing methods. The results are summarized in Tab.\ref{tab:SLR_comparison_more}.

\textbf{}

\begin{figure}[h]
    \centering
    \includegraphics[width=1.0\linewidth]{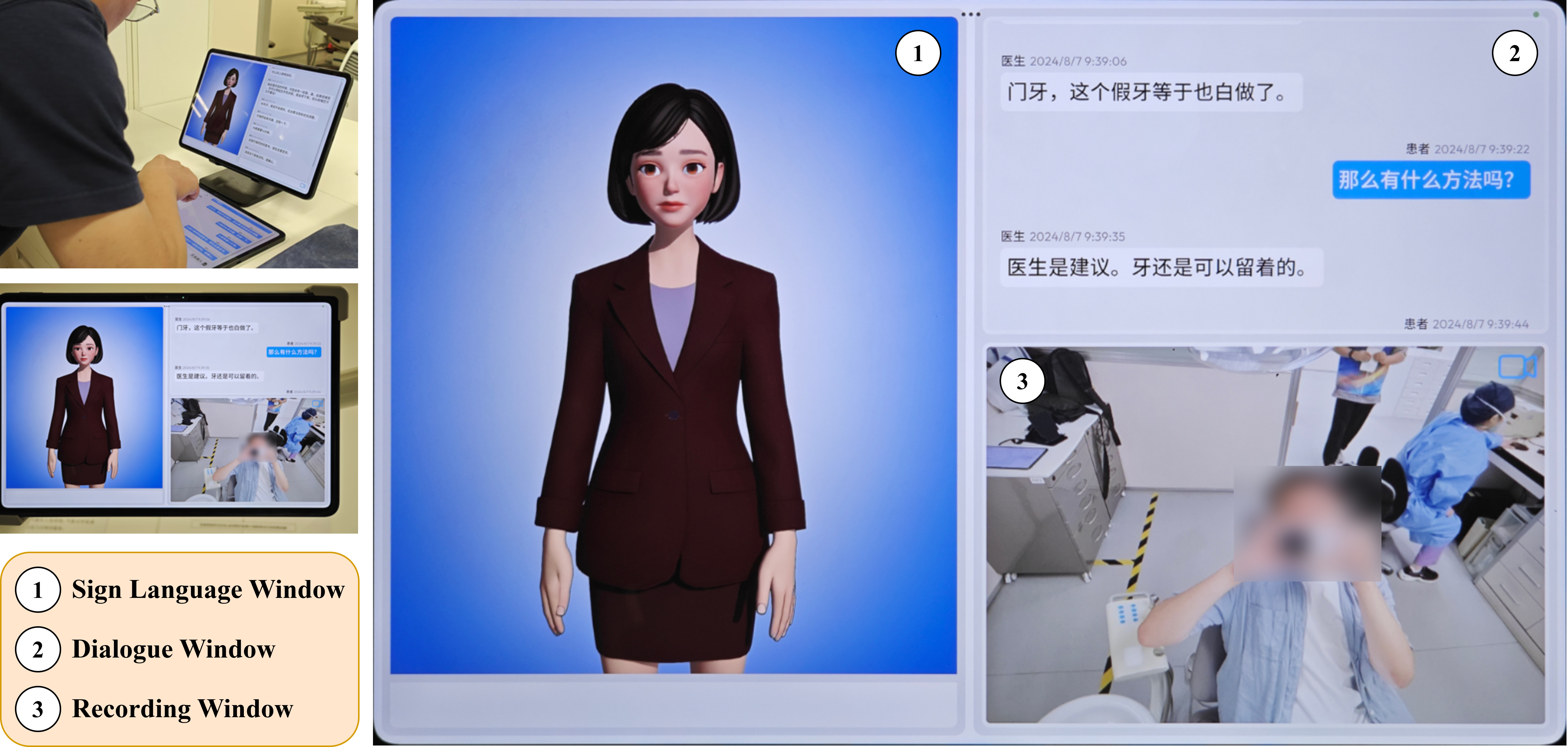}
    \caption{\textbf{\textbf{The demo usage scene in hospital.}}}
    \label{fig:demo}
\end{figure}

\myparagraph{User Study}
We conducted our user study in a local hospital, where Deaf individuals used our demo to communicate with the doctor. The doctor either typed responses or employed speech recognition to generate text for the Deaf patients to read, we also employed sign motion generation via the text. This demo is particularly useful in this context, as Deaf individuals often struggle to see the typing window while receiving dental treatment. An overview of the demo is shown in Fig.\ref{fig:demo}. After using the demo, we asked users to rate the software, as well as the quality of our dataset and inference results, as illustrated in Fig.\ref{fig:user_study}. The results presented in the Experiment section demonstrate that our solution is well-received by the Deaf community.

\subsection{Limitation and Broader Impact}

\myparagraph{Limitation.}
While our SCOPE framework and dataset represent significant advancements in sign language recognition and translation, several limitations must be considered. First, although the SCOPE dataset covers a wide range of scenarios and includes diverse participants, there are still many professional contexts that need to be addressed, such as emergencies (fire, police), insurance consultancy, pet medical treatment, and transportation navigation for the Deaf. Future dataset collection should explore these scenarios.
Second, while SCOPE can incorporate contextual information in sign language processing, excessively long contexts may have minimal impact on current sentences. Investigating an attention decay mechanism should be a future research direction.
Finally, regarding speed normalization, although input image sizes are scaled using our iris normalization, variations in sign speed among signers can complicate the recognition network's ability to process sign frames effectively. Future explorations should focus on addressing sign speed variations.

\myparagraph{Broader Impact.}
The development of the SCOPE framework and dataset offers significant benefits to both the Deaf community and the research field. By enhancing the accuracy and context-awareness of sign language recognition, our work fosters better communication between Deaf and hearing individuals, promoting inclusiveness. Additionally, the SCOPE dataset serves as a valuable resource for developing educational tools that teach sign language and raise awareness in the broader community.
Our real-time sign language translation application is particularly advantageous in healthcare settings, where effective communication is essential for improving care quality for Deaf patients. Furthermore, the insights gained from our research can contribute to advancements in areas such as gesture recognition, virtual reality, and augmented reality.
By open-sourcing our dataset and code, we aim to stimulate further research in sign language processing, leading to innovations that benefit the community. In summary, our work has the potential to drive positive change and innovation, ultimately contributing to a more inclusive and accessible society.

\end{document}